%% file: root.tex
\documentclass[letterpaper, 10 pt, conference]{ieeeconf}  

\usepackage{xcolor} 
\usepackage{caption}
\usepackage{url}
\usepackage{amsmath}
\usepackage{amssymb}
\usepackage{dsfont}
\usepackage{graphicx}
\usepackage{subcaption}
\usepackage{booktabs}
\usepackage{adjustbox}
\usepackage{lipsum}
\definecolor{cvprblue}{rgb}{0.21,0.49,0.74}
\usepackage[pagebackref,breaklinks,colorlinks,citecolor=cvprblue]{hyperref}
\usepackage{cite}

\IEEEoverridecommandlockouts                              

\title{\LARGE \bf
PIRATR: Parametric Object Inference for Robotic Applications with Transformers in 3D Point Clouds
}
\author{Michael Schwingshackl$^{*}$, Fabio F. Oberweger$^{*}$, Mario Niedermeyer, Huemer Johannes, Markus Murschitz \\
AIT Austrian Institute of Technology\\Center for Vision, Automation \& Control}

\begin{document}

\twocolumn[{
\renewcommand\twocolumn[1][]{#1} 
\maketitle
\begin{center}
  \includegraphics[width=\textwidth]{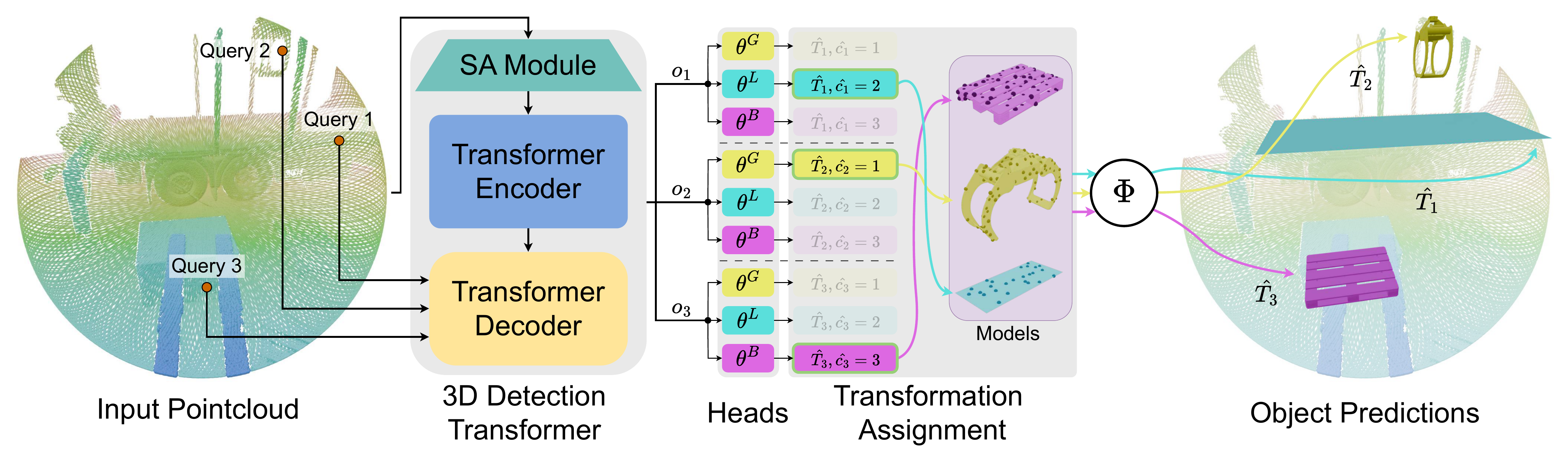}
  {\captionsetup{hypcap=false}
  \captionof{figure}{
    PIRATR is an end-to-end trainable model that takes a point cloud as input, applies farthest point sampling to generate point queries, and encodes them into output embeddings using 3DETR~\cite{misra2021end}. 
    Class-specific heads predict parametric objects corresponding to the gripper, loading platform, and pallet. 
    During training, a geometry-aware matcher~\cite{oberweger2025pi3detr} learns to associate embeddings with the correct classes, and at inference, the model directly outputs class-specific configurations, which are applied to predefined meshes to generate the final predictions. 
  }}
  \label{fig:method_overview}
\end{center}
\vspace{1em}
}]

\thispagestyle{empty}
\pagestyle{empty}

\renewcommand{\thefootnote}{\fnsymbol{footnote}}
\setcounter{footnote}{1}
\footnotetext[1]{Equal contribution.}
\makeatother

\renewcommand{\thefootnote}{\arabic{footnote}} 
\setcounter{footnote}{0}

\input{sections/abstract}
\input{sections/introduction}

\input{sections/related_work}
\input{sections/method}

\input{sections/experiments}
\input{sections/conclusion}

\bibliographystyle{IEEEtran}
\bibliography{biblo/main}

\end{document}

%% file: sections/abstract.tex
\begin{abstract}
We present PIRATR, an end-to-end 3D object detection framework for robotic use cases in point clouds. Extending PI3DETR, our method streamlines parametric 3D object detection by jointly estimating multi-class 6-DoF poses and class-specific parametric attributes directly from occlusion-affected point cloud data. This formulation enables not only geometric localization but also the estimation of task-relevant properties for parametric objects, such as a gripper’s opening, where the 3D model is adjusted according to simple, predefined rules. The architecture employs modular, class-specific heads, making it straightforward to extend to novel object types without re-designing the pipeline. We validate PIRATR on an automated forklift platform, focusing on three structurally and functionally diverse categories: crane grippers, loading platforms, and pallets. Trained entirely in a synthetic environment, PIRATR generalizes effectively to real outdoor LiDAR scans, achieving a detection mAP of 0.919 without additional fine-tuning. PIRATR establishes a new paradigm of pose-aware, parameterized perception. This bridges the gap between low-level geometric reasoning and actionable world models, paving the way for scalable, simulation-trained perception systems that can be deployed in dynamic robotic environments. Code: \href{https://github.com/swingaxe/piratr}{https://github.com/swingaxe/piratr}

\end{abstract}

%% file: sections/introduction.tex
\section{Introduction}
\label{sec:introduction}
Skilled labor shortages drive the recent automation efforts of working machines such as forklifts, cranes, and wheel loaders. Therefore, autonomous robotic machines are following the footsteps of autonomous vehicles into the outdoors. While autonomous vehicles are usually solely focusing on navigating in their environment, autonomous machines i.e., robots interact with their environment to fulfill their purpose. In practical applications such as automated forklifts, interactions are typically limited to structured objects (e.g., pallets) or other machines (e.g., loading platforms) and require reliable 6-DoF pose estimation. Some objects, like crane grippers, can change their state (e.g., opening and closing), so learning a parametric representation is crucial for understanding and interacting with the environment. This data modality, using LiDAR is exemplified in Fig.~\ref{fig:camera_lidar_annotation}.
LiDAR has shown to be a robust sensor modality yielding accurate 3D information even at high distances that can avoid some of the typical problems of vision-based systems, especially their dependence on lighting conditions. While there exist well-established machine learning systems for multi-class object detection and pose estimation in images and RGB-D data, point cloud based detectors are mostly composed of different computation blocks, often containing non-differentiable classical shape fitting  or prepossessing methods. Also, many require an additional dense intermediate representation (such as bird's eye view or voxels).

To this end, this work's approach is inspired by the end-to-end transformer model for 3D edge detection presented in PI3DETR \cite{oberweger2025pi3detr}, which extends 3DETR \cite{misra2021end} beyond bounding boxes to parametric 3D curves but works only on nearly complete point clouds, which are the result of a multi-view scanning system \cite{staderini2023surface, staderini2024visual}. In contrast, our work accounts for incomplete point clouds caused by occlusions and the characteristics of LiDAR systems and, for the first time, successfully applies this approach in the context of contact interaction scenarios of automated machines. Our method operates directly on 3D point clouds, including occlusion artifacts as commonly observed in off-the-shelf LiDAR scans. Owing to its lean end-to-end design, the framework can be readily extended beyond 6-DoF pose estimation to incorporate object-specific state parameters. We demonstrate this capability through gripper detection and localization with joint estimation of the opening angle. PIRATR is trained exclusively on synthetic data and transfers its detection capabilities to real-world point clouds. Consequently, adapting PIRATR to new applications only requires CAD models of the target objects within the synthetic data generation pipeline.
Our contributions presented in this work are:
\begin{itemize}
    \item PIRATR, a generic, fully differentiable end-to-end network for 3D multi-class, parametric multi-object detection and 6-DoF object pose estimation, without intermediate representations (Sec. \ref{sec:PIRATR}) that operates on LiDAR data containing occlusion artifacts. 
    \item A synthetic data generation approach used to train PIRATR that simulates occlusion and other sensor-specific properties and artifacts (Sec. \ref{sec:synthetic}).
    \item Integration of additional (beyond 6-DoF) object state estimation by adding a class-specific feed-forward network per state parameter, which is demonstrated by gripper opening angle estimation (Sec. \ref{subsubsec:parametrized}).
    \item Thorough analysis of the system's capabilities on synthetic and annotated real-world datasets (Sec. \ref{sec:experiments}).
    \item PIRATR deployment on a forklift to perform fully autonomous loading tasks (see Fig. \ref{fig:crayler}).
\end{itemize}



%% file: sections/related_work.tex
\section{Related work}
\label{sec:related_work}
Object detection in 3D point clouds has been widely studied in autonomous driving and mobile robotics. Deep learning methods mainly follow three paradigms: voxel, point, and bird’s-eye view (BEV). Voxel methods discretize the point cloud into 3D grids for convolutional feature extraction~\cite{huang2024voxel}, but suffer from quantization artifacts and high computational cost on dense LiDAR data. Point-based methods such as PointNet~\cite{qi2017pointnet} and its variants \cite{qi2017pointnet++, qian2022pointnext} operate directly on raw points and learn permutation-invariant features, capturing fine local geometry but scaling poorly in large outdoor scenes. BEV approaches \cite{mohapatra2021bevdetnet, Barrera2021BirdNet+:} instead project point clouds onto a top-down grid. PointPillars \cite{lang2019pointpillars} is a notable example encoding vertical columns into pseudo images for efficient 2D convolution, though discretization again causes information loss along the $z$-axis.

Most existing methods predict coarse 3D bounding boxes. Parametric and primitive-based alternatives instead aim for more detailed representations. Classical approaches use geometric fitting such as RANSAC planes \cite{rs9050433}, cylinder fitting \cite{s21227630}, or primitive decomposition \cite{8977383, li2022ransac}, which yield compact and interpretable models but are sensitive to noise, clutter, and missing data. Recent learning-based methods overcome these issues by directly predicting surfaces, edges, or primitives, as in SED-Net \cite{surface_edge_detection}, feature-preserving reconstruction \cite{rs15123155}, or transformer-based sketch-to-primitive models like Point2Primitive \cite{wang2025point2primitive}. Other works estimate parametric objects that follow predefined rules, including extrusion cylinders \cite{uy-point2cyl-cvpr22}, building wireframes \cite{huang2024pbwr}, or CAD models~\cite{WANG2025103838, wang2025point2primitive}.

\begin{figure}[t]
    \centering
    \includegraphics[width=1\linewidth]{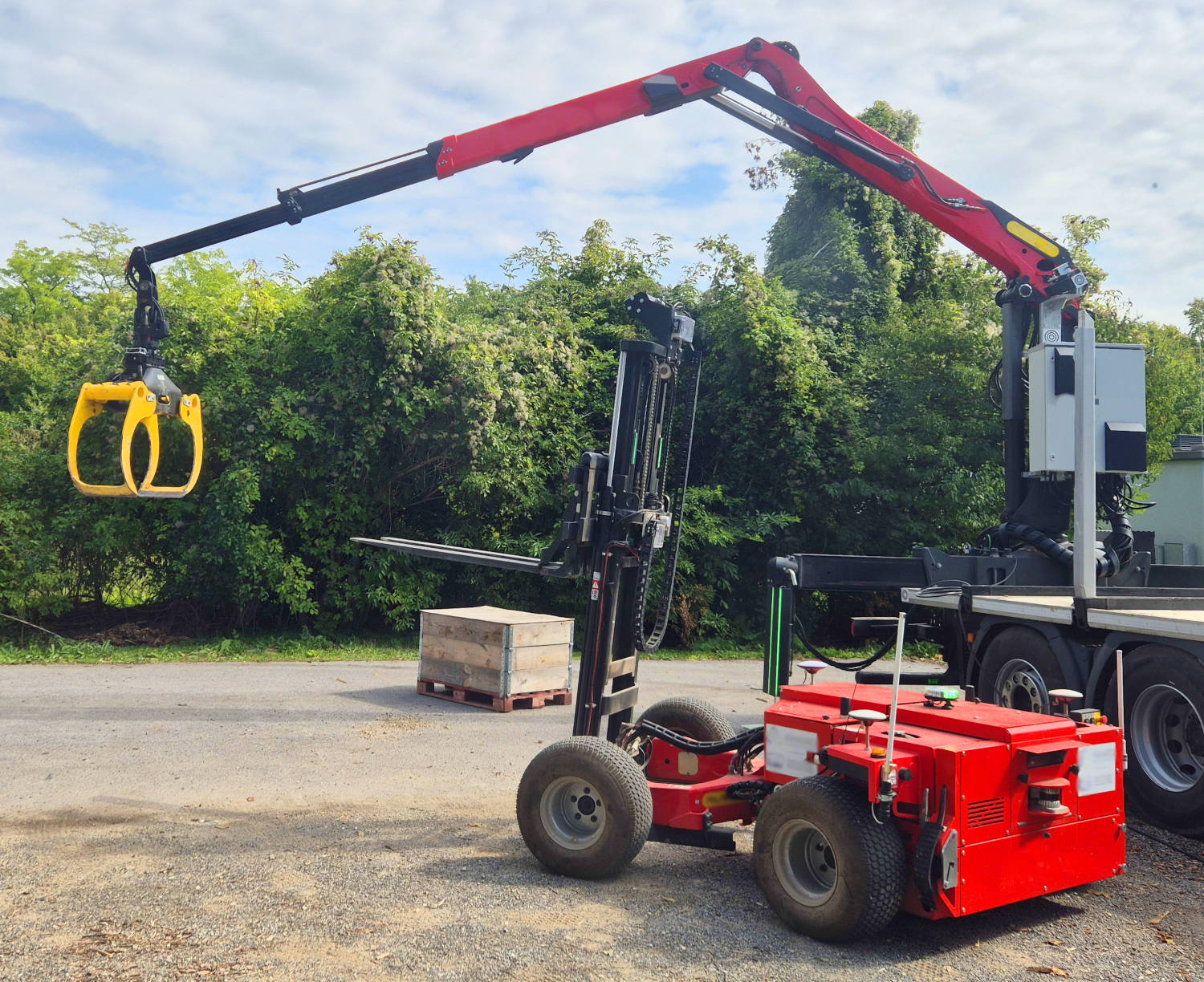}
    \caption{Image of the autonomous forklift operating in an outdoor environment, where our parametric object detection method is tested and deployed. The main detection targets such as the gripper, loading platform and pallets are visible to provide understanding of the perception task and context.}
    \label{fig:crayler}
\end{figure}  
\begin{figure*}
    \centering
    \includegraphics[width=1\linewidth]{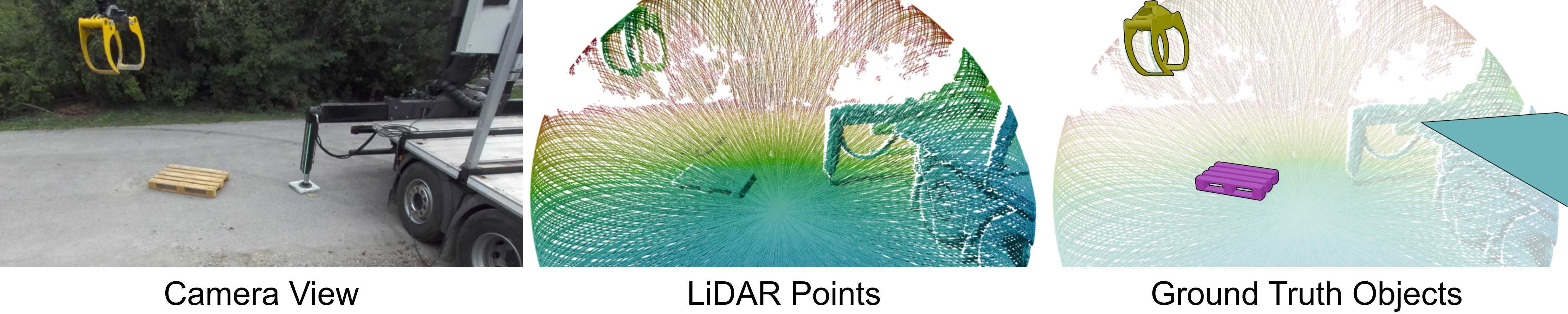}
    \caption{\textbf{Left}: reference image from the forklift-mounted camera. \textbf{Center}: input point cloud captured with a Livox Mid70 LiDAR. \textbf{Right}: 3D annotations of the supervision targets: gripper, loading platform, and pallets.}
    \label{fig:camera_lidar_annotation}
\end{figure*}
PI3DETR~\cite{oberweger2025pi3detr} combines point-based feature extraction with a 3D detection transformer~\cite{carion2020end,misra2021end} architecture to directly predict 3D parametric edges using curve primitives such as lines, Bézier curves, arcs, and circles.

Recent pallet detection methods rely on adaptive Gaussian point feature histograms~\cite{s22208019}, although most approaches operate in 2D or use RGB-D data~\cite{beleznai2024pallet, huemer2025adapt}. Loading platforms or flatbeds are typically represented as planar surfaces, and plane fitting techniques such as RANSAC or robust PCA are commonly employed~\cite{huemer2025adapt}. Other approaches leverage flatbed edge segmentation to estimate usable surface regions~\cite{zou2022fast}.

Voxel-, point-, and BEV-based methods provide robust 3D detection but lack the geometric detail needed for precise manipulation. Parametric and primitive-based approaches yield interpretable representations yet remain limited to simple shapes or narrow tasks. Our work extends the PI3DETR architecture to 3D object detection for robotic outdoor manipulation scenarios. Instead of predicting parametric curves, we predict the 6-DoF pose and individual parameters of pallets, loading platforms, and crane grippers using an end-to-end model, enabling direct deployment in robotic applications.

%% file: sections/method.tex
\section{Method}\label{sec:method}

\subsection{Synthetic Data Generation}
\label{sec:synthetic}
To avoid the cost and effort of collecting annotated data with heavy machinery, we generate synthetic training samples in Blender. This enables rapid creation of diverse scenarios without manual intervention. In total, $5$k samples were produced and split into train ($90\%$), test ($10\%$), and validation ($10\%$) sets.
\subsubsection{Scene generation}
\begin{figure}[t!]
    \centering
    \begin{subfigure}[b]{1\linewidth}
        \centering
        \includegraphics[width=\linewidth]{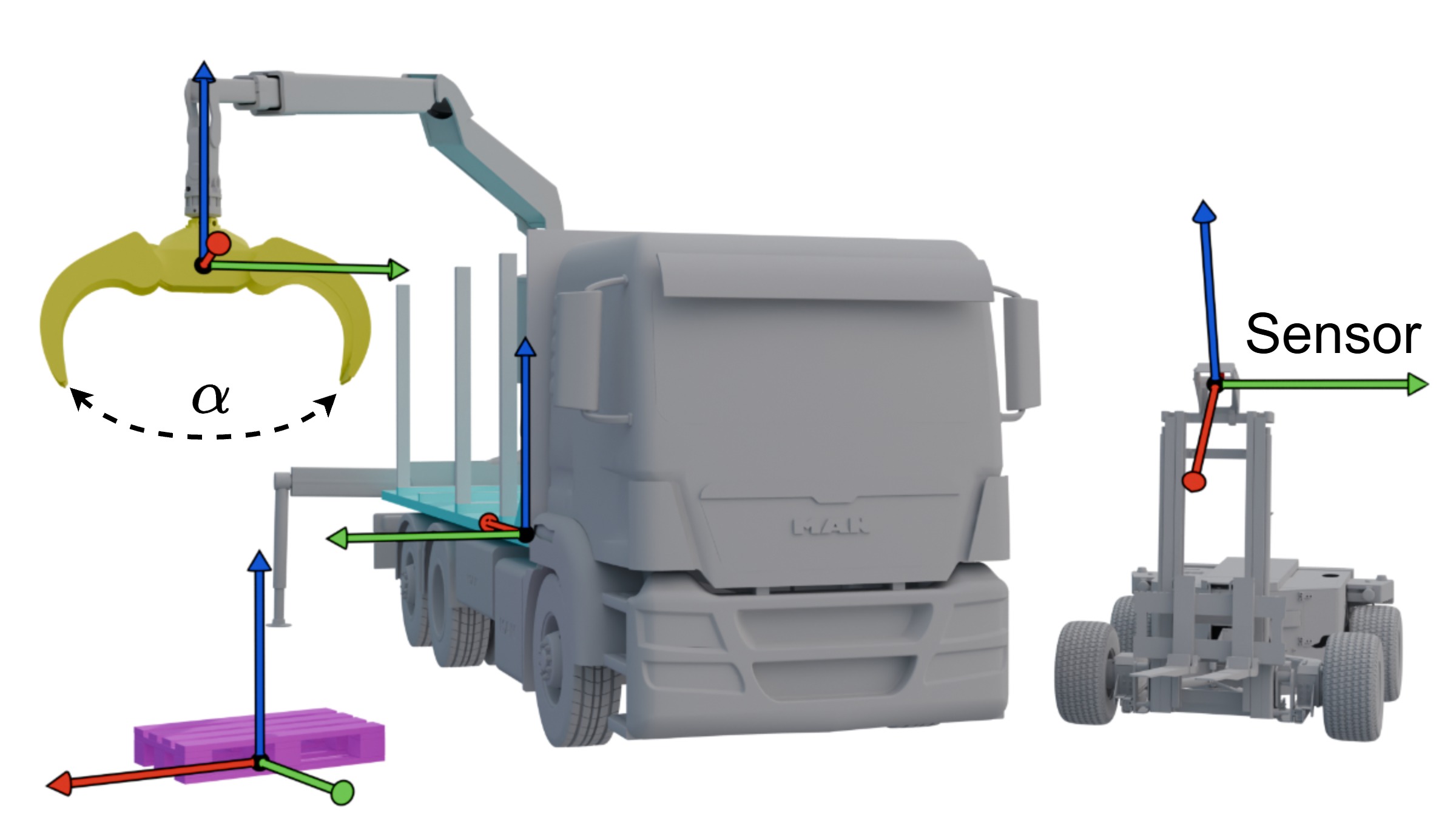}
         \caption{CAD models of the truck-mounted loading crane, forklift, and pallets. Supervision targets: gripper, loading platform, and pallets. Target reference coordinate frames are visualized as well as gripper opening value $\alpha$. The simulated LiDAR sensor is mounted on the forklift mast, matching the real-world configuration.}
        \label{fig:3d_models}
    \end{subfigure}
     \vspace{0.5em} 
    \begin{subfigure}[b]{1\linewidth}
        \centering
        \includegraphics[width=\linewidth]{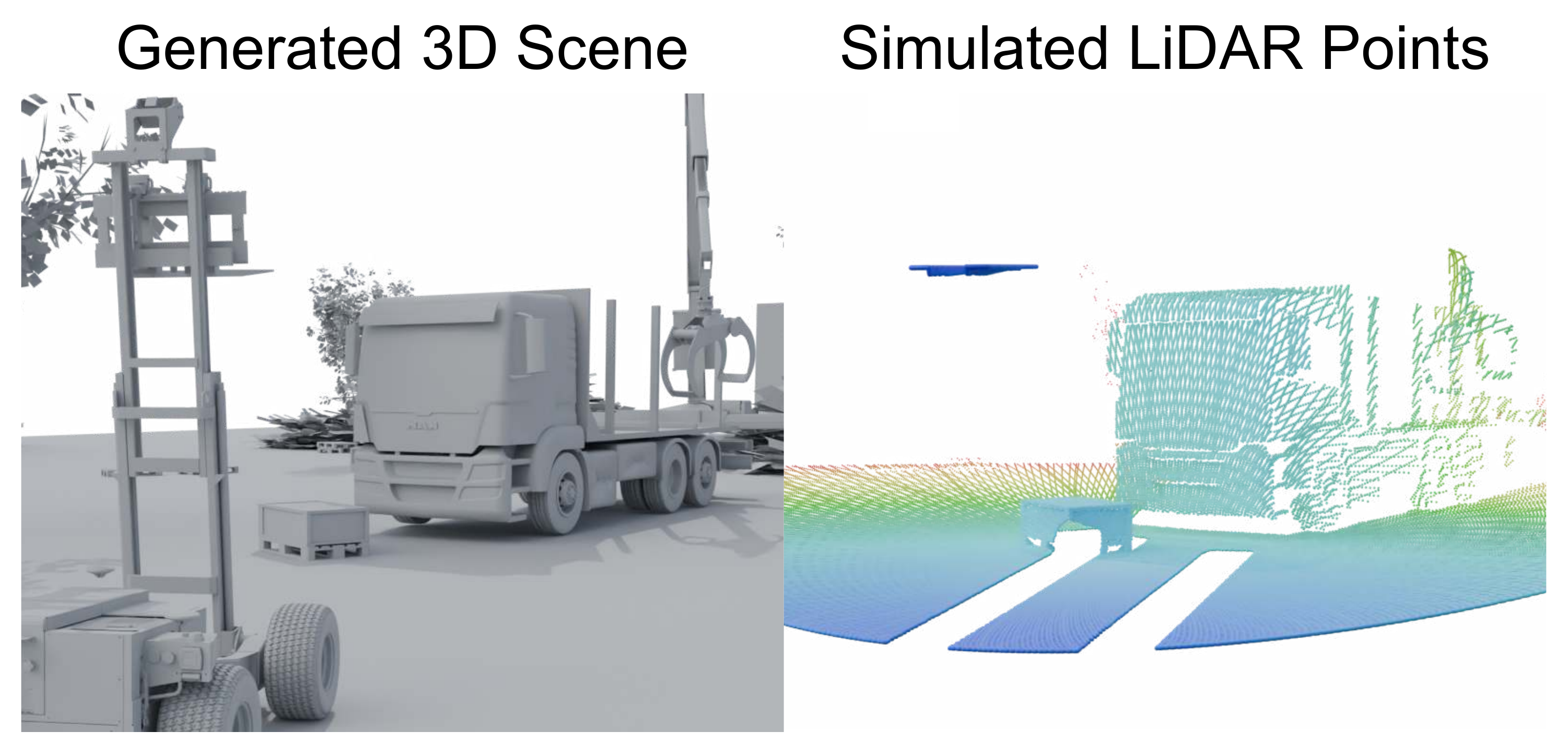}
         \caption{\textbf{Left}: Synthetic training scene generated in Blender with a forklift, truck-mounted crane, pallets, and surrounding clutter/vegetation. \textbf{Right}: Simulated Livox Mid-70 point cloud of the same scene.}
        \label{fig:synthetic_data}
    \end{subfigure}
    \caption{(a) CAD models and supervision targets. (b) Synthetic training scene and corresponding simulated point cloud.}
    \label{fig:synthetic_overview}
\end{figure}

Closing the sim-to-real gap is challenging because simulation hardly reproduces the fine-grained, high-entropy complexity of the real world. To mitigate this, we use accurate CAD models of the forklift and truck-mounted loading crane (omitting only small details such as cables and pistons), as visible in Fig. \ref{fig:3d_models}, combined with extensive domain randomization. The truck-mounted loading crane aligned at the global scene origin. Its gripper, the loading platform, and surrounding pallets form the core of our supervision setup, as their poses (and the gripper’s opening state) are used as training labels following the conventions in Sec. \ref{sec:params_ground_truth}. To randomize scenes, we sample object placements from carefully chosen ranges that reflect realistic configurations. First, the forklift is placed at random positions along a circle of radius $5$–$16$ m around the crane, oriented either toward one of eight predefined points of interest on the truck or away from it, which encourages the model to not assume the truck is always visible. The simulated LiDAR (Sec. \ref{sec:lidar_sim}) is mounted on the forklift's mast, as visible in Figure \ref{fig:3d_models}, so sensor height and tilt vary naturally with mast motion. Similar to the forklift, the crane's gripper is randomly sampled along a circle of radius $3.5$–$8$ m centered at the backside of the truck, with randomized height ($0.5$–$4.5$ m), orientation, and opening angle $\alpha$. Realistic distribution of pallets is achieved using Poisson-disk sampling modulated with fractal Perlin noise to create clusters, while preventing overlaps by keeping proximity to the truck, gripper, and forklift larger than $1.5$m. Pallets may contain either boxes or other pallets, allowing stacks up to two objects high. Since this forklift is intended to work in outdoor terrain, different-sized trees and bushes are scattered with Poisson-disk sampling under a minimum spacing of $1$ m. To simulate urban structure, flat wall meshes are randomly placed around the scene. Fig. \ref{fig:synthetic_data} illustrates an example scene, including the simulated point cloud sampled from the forklift perspective.

\subsubsection{LiDAR simulation}
\label{sec:lidar_sim}

In deep learning with point cloud data, the spatial distribution of points plays a crucial role for both learning stability and downstream performance \cite{guo2020deep}. To ensure realistic input, we replicate the LiDAR scan pattern as closely as possible. Unlike conventional spinning LiDARs, Livox devices employ a non-repetitive scanning mechanism that produces trajectories similar to Lissajous curves, which gradually yield uniform coverage over time. 
To simulate this behavior, we rely on the official Livox-SDK simulation data\footnote{\url{https://github.com/Livox-SDK}}, which provides sequential text files containing timestamp, azimuth, and elevation values for over $400$k rays. The direction vectors for each ray are obtained via a spherical-to-Cartesian coordinate transformation and are subsequently used as inputs to the Raycast node in Blender Geometry Nodes. For each timestamp, the node performs a hit test on the scene geometry from the sensor origin along every ray vector, and the intersection points are collected to form a simulated point cloud $P$. The maximum hit distance is set to $25$m. $P$ is subsequently reduced to $32$k points using farthest point sampling. Targets with LiDAR hit counts below a class-specific threshold are excluded from the ground-truth annotations due to occlusions.


\subsubsection{Parametrization and ground truth definition}
\label{sec:params_ground_truth}
Following \cite{oberweger2025pi3detr}, we define ground-truth annotations as follows. 
Let $P$ denote an input point cloud containing $M$ parametric objects. 
Each object instance is indexed by $i \in \{1,\dots,M\}$ and assigned a class label 
$c_i \in C =\{1,2,3\}$, where $1$, $2$, and $3$ correspond to \emph{grippers}, 
\emph{loading platforms}, and \emph{pallets}, respectively. 
The ground-truth targets are parameterized as
\begin{equation}
T_i =
\begin{cases}
[\mathbf{p}_i, \mathbf{q}_i, \alpha_i], & c_i=1, \\
[\mathbf{p}_i, \mathbf{q}_i], & c_i \in \{2,3\},
\end{cases}
\end{equation}
where $\mathbf{p}_i \in \mathbb{R}^3$ denotes the object position, 
$\mathbf{q}_i \in \mathbb{R}^4$ a unit quaternion encoding its orientation, 
and $\alpha_i \in \mathbb{R}^+$ the opening angle of the gripper.

The interpretation of $\mathbf{p}_i$ is class-dependent. 
For grippers ($c_i=1$), we use the center of mass, as they can translate and rotate freely. For loading platforms ($c_i=2$), we select the front-right corner (facing the driver cabin), since the front/back distinction is semantically relevant and this corner is geometrically well-defined in the point cloud. For pallets ($c_i=3$), we use the bottom center point, reflecting the fact that pallet rotations are constrained by the supporting surface. Moreover, both grippers and pallets are treated as invariant under a $180^\circ$ rotation around the vertical $z$-axis.

As meshes are available for all object classes, explicit dimension regression is not required. 
Each class mesh $\mathcal{M}^c$, with $c \in C$, can be instantiated in the scene using parameters $T_i$, which specify the object’s pose and, in the case of grippers, its state parameter (opening angle). To obtain a compact point-based representation, we associate each mesh with a predefined set of 64 surface points. We formalize this process with a class-conditional mapping
\begin{align}
    \Phi : (\mathcal{M}^{c_i}, T_i) \mapsto \mathbb{R}^{64 \times 3},
\end{align}
which applies the configuration $T_i$ to the mesh $\mathcal{M}^{c_i}$ of instance $i$ and outputs its corresponding 64-point representation. For brevity, we also use $\Phi$ to denote the mapping that applies the configuration to the mesh alone, without extracting the point-based representation.

Finally, all point clouds, meshes, and position parameters are normalized with respect to the longest axis of the input cloud, ensuring that all points lie within $[-1,1]$. The gripper opening angle $\alpha$ is likewise scaled to the range $[-1,1]$ for consistency with the other parameters. Orientations are uniformly represented using unit quaternions.
\begin{figure*}
\centering
    \includegraphics[width=1\linewidth]{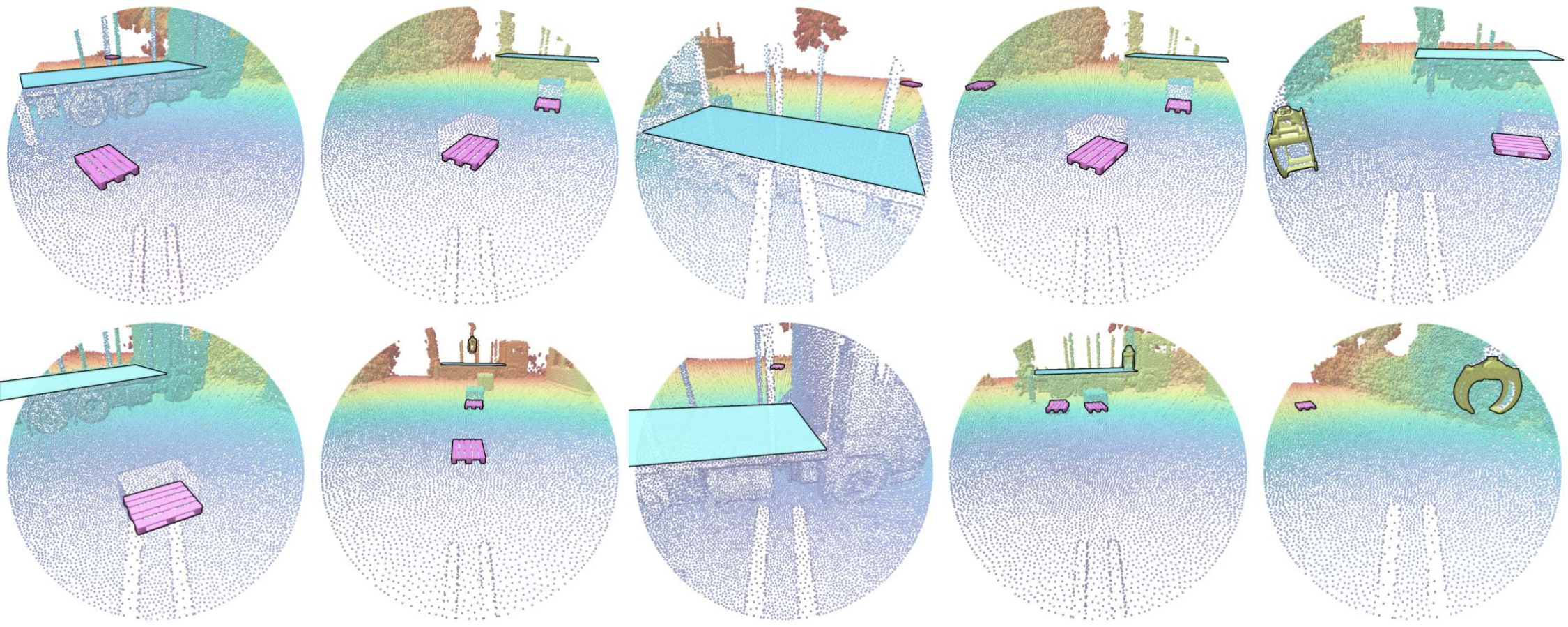}
    \caption{Qualitative synthetic-to-real prediction of PIRATR, which is trained solely on synthetic data and evaluated on real scans. Predicted classes: \textbf{gripper} (yellow), \textbf{loading platforms} (cyan), and \textbf{pallets} (magenta).}
    \label{fig:qualitative_s2r}
\end{figure*}

\subsection{PIRATR Overview}
\label{sec:PIRATR}
Our method builds upon PI3DETR \cite{oberweger2025pi3detr}, which in turn extends the end-to-end 3D object detection framework of \cite{misra2021end}. Our method enhances these foundations by enabling the detection of differently parameterized objects, even under occlusion, 
for robotic applications in point clouds. Given the strong empirical performance of PI3DETR in its original domain, we directly adopt its model core.

In brief, an input point cloud $P$ is first processed by a set-aggregation module (SAModule) \cite{qi2017pointnet++}, which applies farthest point sampling to down-sample the point cloud and extract 
$d$-dimensional local features. These features are passed to a transformer that operates on 
$K$ non-parametric query embeddings. The queries are initialized from farthest point sampled locations $\{q_j\}_{j=1}^K$ and 
encoded via sinusoidal positional embeddings \cite{vaswani2017attention}, yielding a set of output embeddings $\{o_j\}_{j=1}^K$.

Our contribution begins at this stage. Rather than revisiting the core architecture, we refer readers to 
\cite{oberweger2025pi3detr} for a comprehensive description. Here, we focus on the novel extensions introduced by PIRATR to adapt PI3DETR for robotic manipulation scenarios.

\subsubsection{Prediction feed-forward networks (FFN)}
\label{subsubsec:parametrized}
Following \cite{oberweger2025pi3detr}, we define separate prediction heads for each class of parameterized object. Each head is implemented as a feed-forward network (FFN) with parameters $\theta^{(\cdot)}$ and takes as input an output embedding $o_j$ from the 
model core.
For grippers, $\theta^{\text{G}}$ maps $o_j$ to $\hat{T}_j = [\hat{\mathbf{p}}_j, \hat{\mathbf{q}}_j, \hat{\alpha}_j]$, where $\hat{\mathbf{p}}_j \in \mathbb{R}^3$ is the predicted position, $\hat{\mathbf{q}}_j \in \mathbb{R}^4$ a unit quaternion, and $\hat{\alpha}_j \in \mathbb{R}$ the gripper opening angle. Unlike \cite{oberweger2025pi3detr}, we do not employ a separate scalar head. Instead, the FFN jointly predicts all parameters.
The same structure is used for loading platforms ($\theta^{L}$) and pallets ($\theta^{B}$), with outputs $\hat{T}_j = [\hat{\mathbf{p}}_j, \hat{\mathbf{q}}_j]$. Here, the semantic meaning of $\hat{\mathbf{p}}_j$ differs by class, as defined in Sec. \ref{sec:params_ground_truth}. This class-dependent, multi-head setup follows \cite{oberweger2025pi3detr} and is motivated by the differing positional semantics and geometric structures across object categories. 
For clarity of presentation, $\hat{\mathbf{p}}_j$ is described as a directly predicted position. In practice, it is predicted as an offset from the corresponding query point $\mathbf{q}_j$, such that $\hat{\mathbf{p}}_j = \hat{\Delta}_j + \mathbf{q}_j$.
Finally, a classification head $\theta^{\text{cls}}$ maps $o_j$ to 
$\hat{\pi}_j \in [0,1]^4$, representing class probabilities over 
\{\textit{no-object}, \textit{gripper}, \textit{loading platform}, \textit{pallet}\}, 
which determine the object type at inference. As in \cite{oberweger2025pi3detr}, PIRATR predicts parameters for all classes per query, while the geometry-aware matcher ensures that gradients are assigned only to the ground-truth head during training.

\subsubsection{Geometry-aware matching}
We introduce a geometry-aware matching strategy that accounts for object symmetries when aligning predictions to ground truth.  
Let $\mathcal{L}_\text{quat}^\mathcal{S}$ denote the unit quaternion symmetry loss with respect to a symmetry set $\mathcal{S}$, defined as
\begin{align}
    \mathcal{L}_\text{quat}^\mathcal{S}(\hat{\mathbf{q}}, \mathbf{q})
    = \min_{\mathbf{q}^\prime \in \mathcal{E}_\mathcal{S}(\hat{\mathbf{q}})} 
      \ell_1(\mathbf{q}^\prime, \mathbf{q}),
\end{align}
where $\mathcal{E}_\mathcal{S}(\mathbf{q})$ denotes the set of quaternions obtained by applying the symmetries in $\mathcal{S}$ to $\mathbf{q}$.
For instance, let $r^\text{z}(\mathbf{q}) = \mathbf{q} \otimes [0,\,0,\,0,\,1]$ denote a $180^\circ$ rotation around the $z$-axis.  
Then, under the symmetries $\{\pm 1, \pm r^\text{z}\}$, we obtain
\begin{align}
\mathcal{E}_{\{\pm1,\pm r^\text{z}\}}(\mathbf{q}) = \{\mathbf{q},\ -\mathbf{q},\ r^\text{z}(\mathbf{q}),\ -r^\text{z}(\mathbf{q})\}.
\end{align}

We formulate the matching problem as a cost-minimal bipartite assignment between the $M$ ground-truth targets and the $K$ model predictions.  
Each output embedding $o_j \in \{o_j\}_{j=1}^K$ generates parameters for all three object types, yielding $3K$ curve hypotheses in total.  
For a prediction $j \in \{1,\dots,K\}$ and target $i \in \{1,\dots,M\}$, the matching cost is given by
\begin{align}
    \mathcal{C}_\text{match}(j, i) = -\hat{p}_j(c_i) + \mathcal{L}_\text{param}(j, i),
\end{align}
where $\hat{p}_j(c_i)$ denotes the predicted probability of query $j$ belonging to class $c_i$ \cite{carion2020end}, and $\mathcal{L}_\text{param}$ is a geometry-specific parameter loss defined as
\begin{align}
\label{eq:loss_param}
\begin{split}
    \mathcal{L}_\text{param}&(j, i) = \ell_1(\hat{\mathbf{p}}_j, \mathbf{p}_i) + \\[4pt]
    &\begin{cases}
        \mathcal{L}^{\{\pm1,\pm r^\text{z}\}}_\text{quat}(\mathbf{q}_j, \mathbf{q}_i) + \ell_1(\alpha_j, \alpha_i),&  c_i=1, \\[4pt]
        \mathcal{L}^{\{\pm1\}}_\text{quat}(\mathbf{q}_j, \mathbf{q}_i),& c_i = 2, \\[4pt]
        \mathcal{L}^{\{\pm1,\pm r^\text{z}\}}_\text{quat}(\mathbf{q}_j, \mathbf{q}_i),& c_i =3,
    \end{cases}
\end{split}
\end{align}
where $\ell_1(\mathbf{x},\mathbf{y}) = \|\mathbf{x} - \mathbf{y}\|_1$ is the element-wise $\ell_1$-distance.  
This formulation jointly penalizes discrepancies in object position, incorporates quaternion-based symmetry invariances (e.g., $180^\circ$ rotation around the $z$-axis for grippers and pallets), and integrates the gripper opening angle when relevant.  

Following \cite{carion2020end} and \cite{oberweger2025pi3detr}, we determine the optimal assignment $\sigma \in \mathfrak{S}_K$ via the Hungarian algorithm \cite{kuhn1955hungarian}, yielding a minimal-cost permutation of predictions to ground-truth targets.  
The $K-M$ unmatched predictions are assigned to the no-object class ($c_{\sigma(i)}=0$), ensuring consistency with the transformer-based detection framework.

\subsubsection{Loss function}
Let $i \in \{1, \dots, K\}$ denote a ground-truth target from the set including the no-object class, and let $j = \sigma(i) \in \{1, \dots, K\}$ be the prediction matched to $i$.
The total loss for the pair $(j,i)$ is defined as
\begin{align}
\label{eq:loss_total}
\begin{split}
    \mathcal{L}&_\text{total}(j, i) = -\, w_{c_i} \log \hat{\pi}_j  \\[2pt]
    &\quad + \mathds{1}_{\{c_i \neq 0\}} \, \mathcal{L}_\text{param}(j, i) \\[2pt]
    &\quad + \mathds{1}_{\{c_i \neq 0\}} \, 
        \mathcal{L}_\text{CD}\!\left(\Phi(\mathcal{M}^{c_i},\hat{T}_j), \Phi(\mathcal{M}^{c_i}, T_i)\right),
\end{split}
\end{align}
where the first term is the weighted cross-entropy loss between the predicted class distribution $\hat{\pi}_j$ and the ground-truth class $c_i$, the second term is the geometry-aware parameter loss described in Eq. \ref{eq:loss_param}, and the third term is a Chamfer distance (CD) between point sets generated from the predicted and ground-truth configurations.
The CD loss is given by
\begin{align}
\label{eq:chamfer}
\mathcal{L}_\text{CD}(X, Y) 
= \tfrac{1}{|X|} \sum_{\mathbf{x} \in X} \min_{\mathbf{y} \in Y} \ell_2(\mathbf{x},\mathbf{y}) 
+ \tfrac{1}{|Y|} \sum_{\mathbf{y} \in Y} \min_{\mathbf{x} \in X} \ell_2(\mathbf{x},\mathbf{y}),
\end{align}
where $\ell_2(\mathbf{x},\mathbf{y}) = \|\mathbf{x} - \mathbf{y}\|_2$ denotes the Euclidean distance.  
Here, the mapping $\Phi(\mathcal{M}^{c_i}, T_i)$ applies the configuration $T_i$ to the class-specific mesh $\mathcal{M}^{c_i}$ and yields its corresponding point-based representation. The Chamfer loss $\mathcal{L}_\text{CD}$ therefore enforces geometric consistency by comparing the posed mesh of the prediction against that of the ground truth through their transformed point sets, ensuring both accurate placement and correct articulation in 3D space.

If $c_i = 0$ (no-object placeholder), the loss reduces to the cross-entropy term.  
Class weights $w_{c_i}$ are computed following \cite{oberweger2025pi3detr}.

\begin{table}[t]
\centering
\begin{subtable}{\linewidth}
\centering
\begin{adjustbox}{max width=\linewidth}
\begin{tabular}{l|ccc}
\toprule
\textbf{Metric} & \textbf{Gripper} & \textbf{Loading Platform} & \textbf{Pallet} \\
\midrule
& \multicolumn{3}{c}{\textit{Geometric Quality}}\\
\cmidrule(lr){2-4}
$\ell_2$ [m] $\downarrow$ & 0.10 {\scriptsize ($\pm$ 0.07)} & 0.12 {\scriptsize ($\pm$ 0.09)} & 0.12 {\scriptsize ($\pm$ 0.08)} \\
Geodesic [$^\circ$] $\downarrow$ & 4.90 {\scriptsize ($\pm$ 4.42)} & 0.92 {\scriptsize ($\pm$ 1.06)} & 12.29 {\scriptsize ($\pm$ 17.08)} \\
Yaw [$^\circ$] $\downarrow$ & 2.37 {\scriptsize ($\pm$ 2.48)} & 0.74 {\scriptsize ($\pm$ 0.95)} & 10.48 {\scriptsize ($\pm$ 14.35)} \\
Opening [$^\circ$] $\downarrow$ & 6.59 {\scriptsize ($\pm$ 10.80)} & -- & -- \\
\midrule
& \multicolumn{3}{c}{\textit{Detection (mAP = 0.982)}} \\
\cmidrule(lr){2-4}
Det. (AP) $\uparrow$ & 0.997 & 0.990 & 0.958 \\
\bottomrule 
\end{tabular}
\end{adjustbox}
\caption{Synthetic test set.}
\label{tab:main_synth}
\end{subtable}

\vspace{0.8em} 

\begin{subtable}{\linewidth}
\centering
\begin{adjustbox}{max width=\linewidth}
\begin{tabular}{l|ccc}
\toprule
\textbf{Metric} & \textbf{Gripper} & \textbf{Loading Platform} & \textbf{Pallet} \\
\midrule
& \multicolumn{3}{c}{\textit{Geometric Quality}}\\
\cmidrule(lr){2-4}
$\ell_2$ [m] $\downarrow$ & 0.13 {\scriptsize ($\pm$ 0.10)} & 0.24 {\scriptsize ($\pm$ 0.18)} & 0.14 {\scriptsize ($\pm$ 0.07)} \\
Geodesic [$^\circ$] $\downarrow$ & 6.37 {\scriptsize ($\pm$ 4.26)} & 1.57 {\scriptsize ($\pm$ 0.73)} & 15.18 {\scriptsize ($\pm$ 22.43)} \\
Yaw [$^\circ$] $\downarrow$ & 2.46 {\scriptsize ($\pm$ 2.20)} & 1.19 {\scriptsize ($\pm$ 0.85)} & 12.32 {\scriptsize ($\pm$ 18.02)} \\
Opening [$^\circ$] $\downarrow$ & 6.69 {\scriptsize ($\pm$ 5.11)} & -- & -- \\
\midrule
& \multicolumn{3}{c}{\textit{Detection (mAP = 0.919)}} \\
\cmidrule(lr){2-4}
Det. (AP) $\uparrow$ & 0.962 & 0.988 & 0.805 \\
\bottomrule 
\end{tabular}
\end{adjustbox}
\caption{Real test set (Synthetic-to-real evaluation).}
\label{tab:main_s2r}
\end{subtable}

\caption{Quantitative test set evaluations for geometric errors and detection AP. Values are mean ($\pm$ std.).}
\label{tab:main_combined}
\end{table}
\newpage
\subsubsection{Implementation details}
\paragraph{Model} PIRATR is implemented in PyTorch \cite{paszke2019pytorch} and follows the model configurations of PI3DETR \cite{oberweger2025pi3detr}. The backbone consists of $3$ Transformer encoder layers and $9$ decoder layers \cite{vaswani2017attention}, with a token feature dimension of $768$. The input to the Transformer is a set of $2048$ points encoded by a SAModule \cite{qi2017pointnet++}, together with $128$ sine-embedded query points \cite{misra2021end} used for prediction. Training is initialized from a checkpoint provided by the authors of PI3DETR. We first train for 330 epochs using AdamW \cite{loshchilov2017decoupled} with a learning rate of $10^{-4}$, followed by continued training of 100 epochs with a reduced learning rate of $10^{-5}$. An effective batch size of $132$ is achieved with gradient accumulation, and gradients are clipped at an $\ell_2$-norm of $0.2$. Losses are normalized per object. Training is performed on a single NVIDIA RTX 4090 GPU. Data augmentation includes random point cloud rotations of up to $5^\circ$ around the $x$- and $y$-axes. In addition, with probability $\tfrac{1}{3}$, Gaussian noise with standard deviation sampled from the interval $(0.0, 0.04]$ is added to the normalized points (see Sec.\ref{sec:params_ground_truth}).

\paragraph{Real Data Inference}
\label{sec:real_data_inference}
The autonomous forklift is equipped with a Livox Mid-$70$ sensor operating at a wavelength of $905$ nm with a circular field of view of $70.4^{\circ}$, delivering $10$ packages of $10$k points per second via a ROS2 \cite{macenski2022robot} interface. Points are filtered according to the Livox user manual\footnote{\url{https://www.livoxtech.com/mid-70/downloads} (accessed on 2025-09-14)}, retaining only those with low noise characteristics in terms of intensity and spatial position. Since our aggregated LiDAR scans can contain between $50$k and $400$k points, we first discard all points beyond $25$m and then subsample the remaining points to $32$k using farthest point sampling. This not only aligns the point count and distance range with our synthetic training data but also ensures an even point distribution, mitigating artifacts caused by high-density regions in the raw scans. The filtered points are rotated $180$° around the z-axis to align with Blender conventions used for training and then passed to the network described in Sec. \ref{sec:PIRATR}. The network’s predicted poses for each parametric object are subsequently rotated back -$180$° around the z-axis to match ROS2 conventions and the original point cloud frame.

\begin{figure}[t]
    \centering
    \includegraphics[width=1\linewidth]{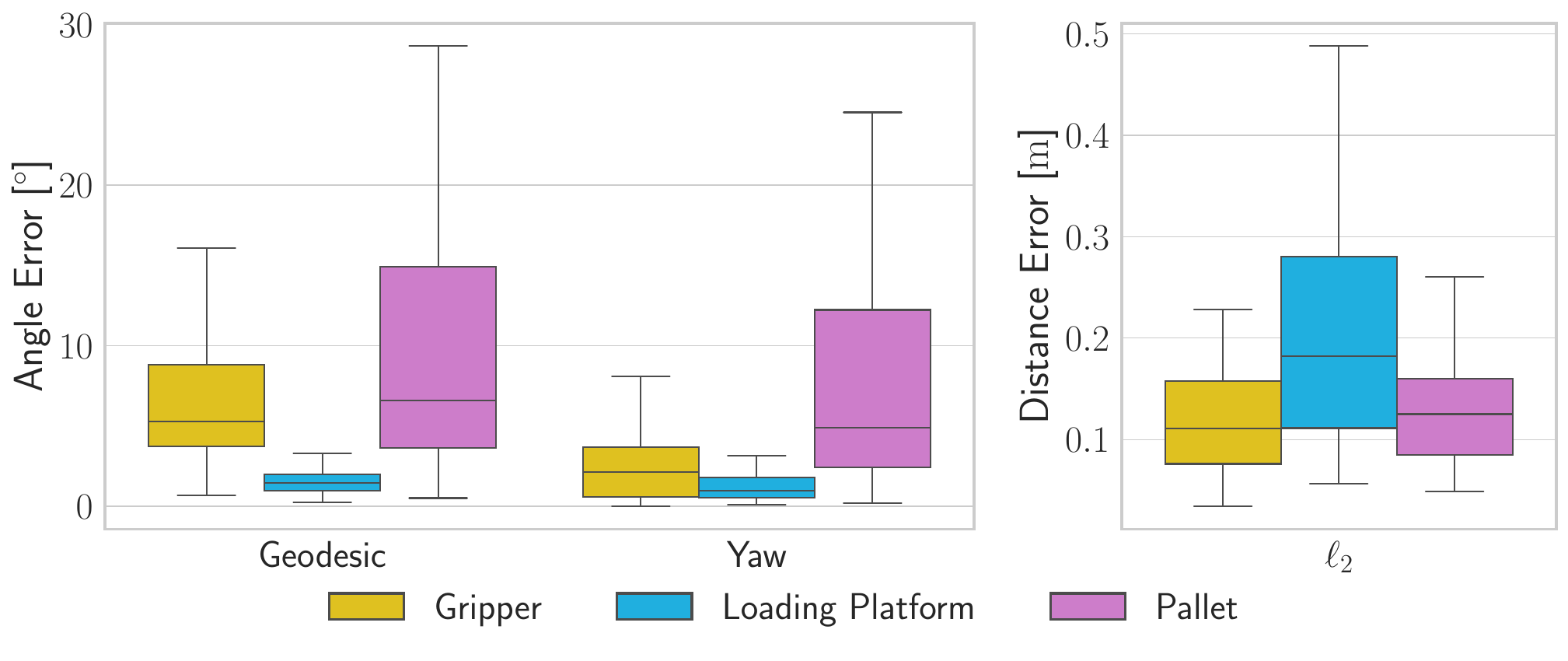}
    \caption{Boxplots of synthetic-to-real evaluation error distributions for angles and distances.}
    \label{fig:boxplot_s2r}
\end{figure}

%% file: sections/experiments.tex
\section{Experiments}\label{sec:experiments}
We evaluate PIRATR on both a synthetically generated dataset and a real-world dataset collected with the Livox Mid-70, which was manually annotated. We present quantitative results on both datasets (Sec. \ref{sec:synth_eval} and Sec. \ref{sec:s2r_eval}), and we include qualitative evaluations as well as robustness tests (Sec. \ref{sec:robustness_eval}) on the real-world dataset. The real-world dataset comprises $73$ scenes containing $34$ grippers, $32$ loading platforms, and $70$ pallets.

To evaluate our model quantitatively, we use two classes of metrics. Geometric quality: we report the $\ell_2$-distance for positional offsets, the geodesic error for full 3D orientation, and the yaw error. For gripper predictions, we additionally measure the error in the predicted opening angle. All angular errors are reported in degrees $(^\circ)$ and the $\ell_2$ is reported in meters (m). Detection accuracy: we report (mean) average precision (AP). A prediction $j$ is considered a match with ground truth $i$ if $\hat{c}_j = c_i$ and $\mathcal{L}_\text{CD}\!\left(\Phi(\mathcal{M}^{c_i}, \hat{T}_j), \Phi(\mathcal{M}^{c_i}, T_i)\right) < 0.00125$. The matches are reused for the computation of the geometric metrics.

\subsection{Syntetic Dataset Evaluation}
\label{sec:synth_eval}
Tab. \ref{tab:main_synth} presents the results on the synthetic test dataset. The proposed method attains a mean average precision (mAP) of $0.982$ and maintains accurate predictions in terms of $\ell_2$-distance, with an average offset below $12$ cm across all classes. The loading platform yields the best scores due to its simpler geometry, while the gripper class is also predicted with high accuracy. The pallet class constitutes the most challenging case, yet the method still achieves an average precision of $0.958$. Rotation estimation remains the most difficult component compared to the other evaluated metrics.

\begin{figure}[t]
    \centering
    \includegraphics[width=1\linewidth]{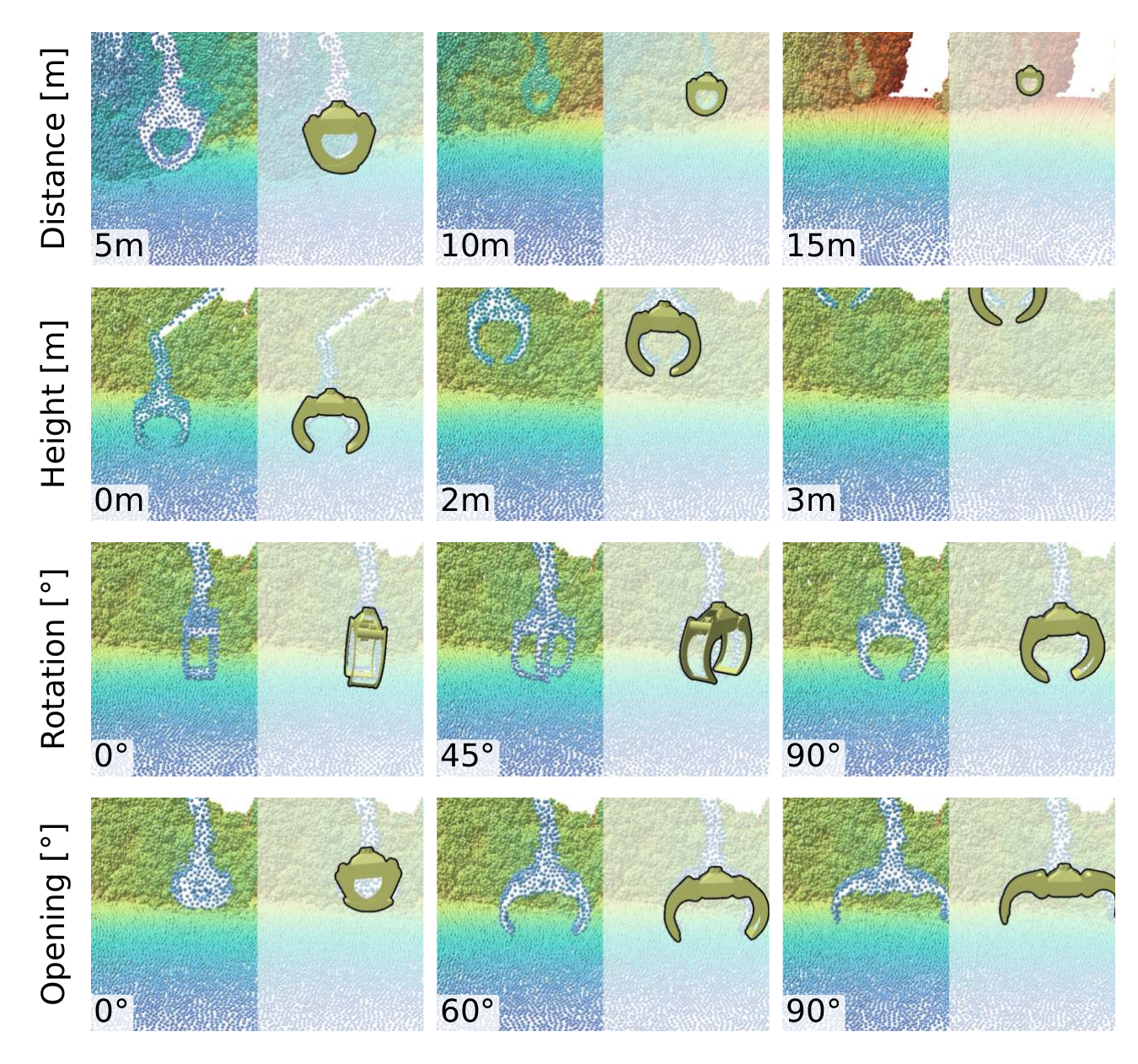}
    \caption{Synthetic-to-real robustness evaluation on gripper predictions.}
    \label{fig:gripper_robustness}
\end{figure}

\begin{figure}
    \centering
    \includegraphics[width=1\linewidth]{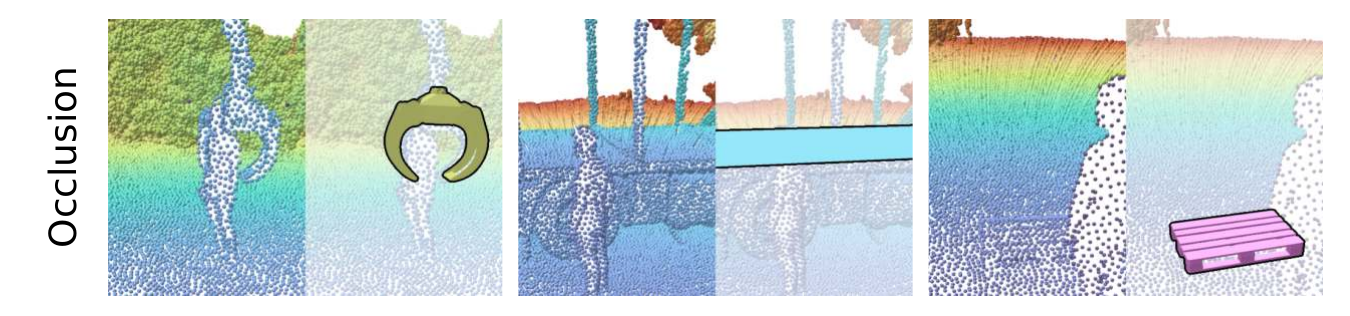}
    \caption{Synthetic-to-real predictions under occlusion scenarios caused by a human standing in front of the objects, shown from left to right: gripper, loading platform, and pallet.}
    \label{fig:occlusion_robustness}
\end{figure}

\begin{figure}
    \centering
    \includegraphics[width=1\linewidth]{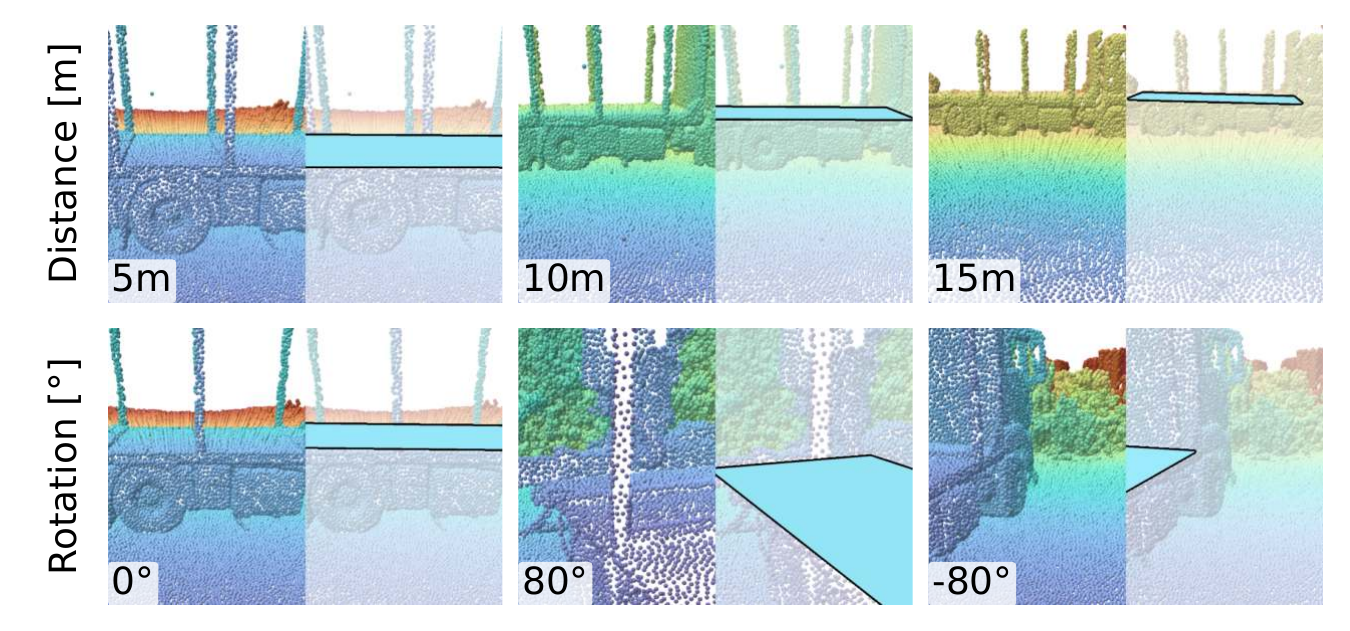}
    \caption{Synthetic-to-real robustness evaluation on loading platform predictions.}
    \label{fig:loading_robustness}
\end{figure}

\begin{figure}
    \centering
    \includegraphics[width=1\linewidth]{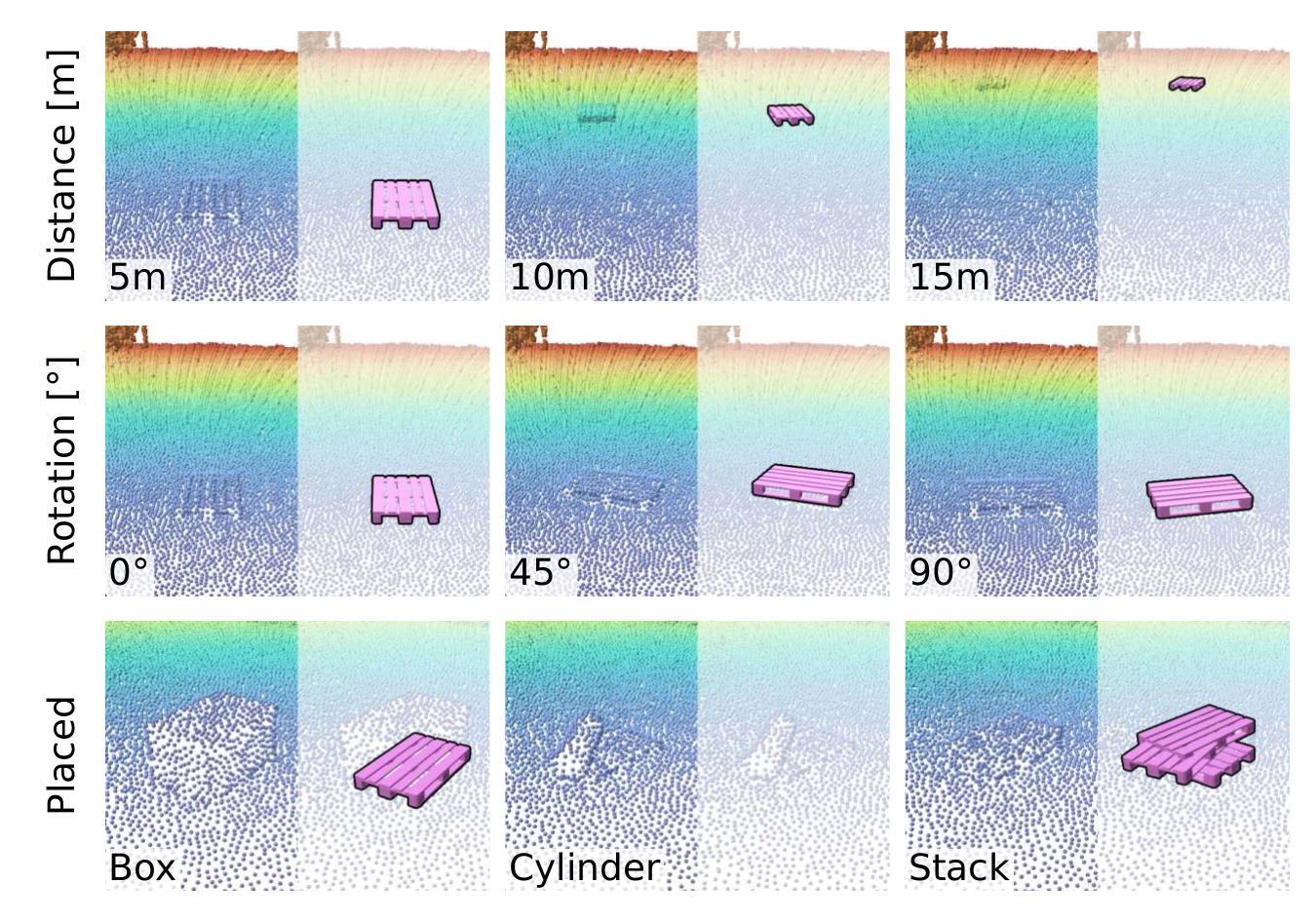}
    \caption{Synthetic-to-real robustness evaluation on pallet predictions.}
    \label{fig:pallet_robustness}
\end{figure}

\subsection{Synthetic-to-Real Evaluation}
\label{sec:s2r_eval}
To assess synthetic-to-real generalization, we evaluate the model trained exclusively on synthetic data on the real world dataset. The results are reported in Tab. \ref{tab:main_s2r}. Compared to the synthetic test results in Tab. \ref{tab:main_synth}, PIRATR achieves performance that remains consistent across most metrics. The largest drop occurs for the pallet class, with an average precision of $0.805$ compared to $0.958$ on synthetic data. This decrease is likely due to the random placement of pallets during synthetic data generation, where some instances appear in vegetation and are still considered valid ground truth if sufficient point coverage is captured. The boxplots in Fig. \ref{fig:boxplot_s2r} complement Tab. \ref{tab:main_s2r} and provide further insight into the error distributions. For all classes, more than $50\%$ of the angle errors are below $10^\circ$, and more than half of the $\ell_2$ errors are below $20$ cm. Overall, the results demonstrate that the proposed model transfers reliably from synthetic to real data, as illustrated in Fig. \ref{fig:qualitative_s2r}.

\subsubsection{Robustness Tests}
\label{sec:robustness_eval}
In this section, we qualitatively evaluate synthetic-to-real robustness across different class-specific scenarios. Fig. \ref{fig:gripper_robustness} and Fig. \ref{fig:loading_robustness} show predictions for grippers and loading platforms under variations in distance, height, rotation, and gripper opening angle. In all figures, the point cloud is shown on the left and the corresponding prediction on the right in a grayed area for clarity. PIRATR demonstrates robust performance for both classes, with the only strong inaccuracies observed in the close-up $80^\circ$ loading platform sample. For pallets, Fig. \ref{fig:pallet_robustness} presents cases where additional objects such as boxes, cylinders, or stacked pallets are placed on top. In these scenarios, pallets are not detected correctly when a cylinder is placed on top, and for stacks the rotation and number of predictions are inaccurate. Cylinders on pallets were not included in the training data. Finally, Fig. \ref{fig:occlusion_robustness} shows tests with a person occluding the target object across all classes. Despite the absence of such cases in training, the method consistently detects the target objects.

\subsubsection{Point Cloud Accumulation \& Runtime}
Since LiDAR point cloud acquisition depends on aggregation time, we evaluate the performance of the synthetically trained model on different point aggregation counts of real point clouds. Tab. \ref{tab:metrics_agg_comparison} reports metrics for $50$k, $200$k, and $400$k points, which are fed into the model after applying the preprocessing described in Sec. \ref{sec:real_data_inference}. The $400$k setting, which is closest to the synthetic data generation setup, delivers the most stable performance across metrics. Nevertheless, all aggregation counts yield decent performance.
From raw input to final predictions, the method requires $218$ms ($\pm 59$ms), $371$ms ($\pm 62$ms), and $598$ms ($\pm 76$ms) for $50$k, $200$k, and $400$k  on a NVIDIA
RTX 4090 GPU, respectively.

\begin{table}
\centering
\begin{adjustbox}{max width=\linewidth}
\begin{tabular}{l|ccc|ccc|ccc}
\toprule
\textbf{Metric} & \multicolumn{3}{c|}{\textbf{Gripper}} & \multicolumn{3}{c|}{\textbf{Loading Platform}} & \multicolumn{3}{c}{\textbf{Pallet}} \\
\cmidrule(lr){2-4} \cmidrule(lr){5-7} \cmidrule(lr){8-10}
&  $50$k & $200$k & $400$k & $50$k & $200$k & $400$k & $50$k & $200$k & $400$k \\
\midrule
& \multicolumn{9}{c}{\textit{Geometric Quality}} \\
\cmidrule(lr){2-10}
$\ell_2$ [m] $\downarrow$ & \textbf{0.13} & \textbf{0.13} & \textbf{0.13} & 0.21 & 0.28 & \textbf{0.24} & \textbf{0.14} & 0.15 & \textbf{0.14} \\
Geodesic [$^\circ$] $\downarrow$ & 9.12 & 7.15 & \textbf{6.37} & 1.71 & \textbf{1.37} & 1.57 & \textbf{12.52} & 17.90 & 15.18 \\
Yaw [$^\circ$] $\downarrow$ & 3.32 & 2.47 & \textbf{2.46} & 1.24 & \textbf{0.96} & 1.19 & \textbf{10.19} & 14.69 & 12.32 \\
Opening [$^\circ$] $\downarrow$ & 10.49 & 8.90 & \textbf{6.69} & -- & -- & -- & -- & -- & -- \\
\midrule
& \multicolumn{9}{c}{\textit{Detection}} \\
\cmidrule(lr){2-10}
Det. (AP) $\uparrow$ & 0.923 & \textbf{0.992} & 0.962 & 0.940 & 0.940 & \textbf{0.988} & \textbf{0.805} & 0.710 & \textbf{0.805} \\
\bottomrule
\end{tabular}
\end{adjustbox}
\caption{Synthetic-to-real geometric and detection metrics for different point cloud accumulation sizes ($50$k, $200$k, $400$k points).}
\label{tab:metrics_agg_comparison}
\end{table}

%% file: sections/conclusion.tex
\section{Conclusion \& Outlook}\label{sec:conclusion}
We presented PIRATR, an end-to-end framework for parametric 3D object detection that jointly estimates multi-class 6-DoF poses and class-specific attributes from partially observed point clouds. PIRATR showed strong synthetic-to-real transfer on challenging outdoor LiDAR scans despite training solely on simulated data. Its practical relevance was demonstrated in an automated forklift operation outdoors. Future work includes extending supported object classes, improving synthetic data realism, reducing training times, and incorporating temporal information for greater robustness.